\documentclass{llncs}
\usepackage[T1]{fontenc}
\usepackage[utf8]{inputenc}
\usepackage{booktabs}
\usepackage{algorithm2e}
\usepackage{amsmath}
\usepackage{amssymb}
\usepackage{graphicx}
\usepackage[unicode=true]
 {hyperref}

\makeatletter

\providecommand{\tabularnewline}{\\}

\usepackage{setspace}
\usepackage{microtype}

\makeatother

\begin{document}

\title{Limited Evaluation Evolutionary Optimization of Large Neural Networks}

\author{Jonas Prellberg \and Oliver Kramer}

\institute{University of Oldenburg, Oldenburg, Germany\\
\email{\textnormal{\{}jonas.prellberg,oliver.kramer\textnormal{\}}@uni-oldenburg.de}}
\maketitle
\begin{abstract}
Stochastic gradient descent is the most prevalent algorithm to train
neural networks. However, other approaches such as evolutionary algorithms
are also applicable to this task. Evolutionary algorithms bring unique
trade-offs that are worth exploring, but computational demands have
so far restricted exploration to small networks with few parameters.
We implement an evolutionary algorithm that executes entirely on the
GPU, which allows to efficiently batch-evaluate a whole population
of networks. Within this framework, we explore the limited evaluation
evolutionary algorithm for neural network training and find that its
batch evaluation idea comes with a large accuracy trade-off. In further
experiments, we explore crossover operators and find that unprincipled
random uniform crossover performs extremely well. Finally, we train
a network with 92k parameters on MNIST using an EA and achieve 97.6\,\%
test accuracy compared to 98\,\% test accuracy on the same network
trained with Adam. Code is available at \texttt{\href{https://github.com/jprellberg/gpuea}{https://github.com/jprellberg/gpuea}}.
\end{abstract}

\section{Introduction}

Stochastic gradient descent (SGD) is the leading approach for neural
network parameter optimization. Significant research effort has lead
to creations such as the Adam \cite{adam} optimizer, Batch Normalization
\cite{ioffe2015batch} or advantageous parameter initializations \cite{pmlr-v9-glorot10a},
all of which improve upon the standard SGD training process. Furthermore,
efficient libraries with automatic differentiation and GPU support
are readily available. It is therefore unsurprising that SGD outperforms
all other approaches to neural network training. Still, in this paper
we want to examine evolutionary algorithms (EA) for this task.

EAs are powerful black-box function optimizers and one prominent advantage
is that they do not need gradient information. While neural networks
are usually built so that they are differentiable, this restriction
can be lifted when training with EAs. For example, this would allow
the direct training of neural networks with binary weights for deployment
in low-power embedded devices. Furthermore, the loss function does
not need to be differentiable so that it becomes possible to optimize
for more complex metrics.

With growing computational resources and algorithmic advances, it
is becoming feasible to optimize large, directly encoded neural networks
with EAs. Recently, the limited evaluation evolutionary algorithm
(LEEA) \cite{Morse:2016:SEO:2908812.2908916} has been introduced,
which saves computation by performing the fitness evaluation on small
batches of data and smoothing the resulting noise with a fitness inheritance
scheme. We create a LEEA implementation that executes entirely on
a GPU to facilitate extensive experimentation. The GPU implementation
avoids memory bandwidth bottlenecks, reduces latency and, most importantly,
allows to efficiently batch the evaluation of multiple network instances
with different parameters into a single operation.

Using this framework, we highlight a trade-off between batch size
and achievable accuracy and also find the proposed fitness inheritance
scheme to be detrimental. Instead, we show how the LEEA can profit
from low selective pressure when using small batch sizes. Despite
the problems discussed in literature about crossover and neural networks
\cite{GARCIAPEDRAJAS2006514,nonredundantcoding}, we see that basic
uniform and arithmetic crossover perform well when paired with an
appropriately tuned mutation operator. Finally, we apply the lessons
learned to train a neural network with 92k parameters on MNIST using
an EA and achieve 97.6\,\% test accuracy. In comparison, training
with Adam results in 98\,\% test accuracy. (The network is limited
by its size and architecture and cannot achieve state-of-the-art results.)

The remainder of this paper is structured as follows: Section~\ref{sec:Related-Work}
presents related work on the application of EAs to neural network
training. In Section~\ref{sec:Method}, we present our EA in detail
and explain the advantages of running it on a GPU. Section~\ref{sec:Experiments}
covers all experiments and contains the main results of this work.
Finally, we conclude the paper in Section~\ref{sec:Conclusion}.

\section{Related Work\label{sec:Related-Work}}

Morse et al. \cite{Morse:2016:SEO:2908812.2908916} introduced the
limited evaluation (LE) evolutionary algorithm for neural network
training. It is a modified generational EA, which picks a small batch
of training examples at the beginning of every generation and uses
it to evaluate the population of neural networks. This idea is conceptually
very similar to SGD, which also uses a batch of data for each step.
Performing the fitness evaluation on small batches instead of the
complete training set massively reduces the required computation,
but it also introduces noise into the fitness evaluation. The second
component of the LEEA is therefore a fitness inheritance scheme that
combines past fitness evaluation results. The algorithm is tested
with networks of up to 1500 parameters and achieves results comparable
to SGD on small datasets.

Baioletti et al. \cite{10.1007/978-3-319-72926-8_33} pick up the
LE idea but replace the evolutionary algorithm with differential evolution
(DE), which is a very successful optimizer for continuous parameter
spaces \cite{Das2016}. The largest network they experiment with employs
7000 parameters. However, there is still a rather large performance
gap on the MNIST dataset between their best performing DE algorithm
at 85\,\% accuracy and a standard SGD training at 92\,\% accuracy.

Yaman et al. \cite{yaman2018limited} combine the concepts LE, DE
and cooperative co-evolution. They consider the pre-synaptic weights
of a single neuron a component and evolve many populations of such
components in parallel. Complete solutions are created by combining
components from different populations to a network. Using this approach,
they are able to optimize networks of up to 28k parameters.

Zhang et al. \cite{DBLP:journals/corr/abs-1712-06564} explore neural
network training with a natural evolution strategy. This algorithm
starts with an initial parameter vector $\theta$ and creates many
so-called pseudo-offspring parameter vectors by adding random noise
to $\theta$. The fitness of all pseudo-offspring is evaluated and
used to estimate the gradient at $\theta$. Finally, this gradient
approximation is fed to SGD or another optimizer such as Adam to modify
$\theta$. Using this approach, they achieve 99\,\% accuracy on MNIST
with 50k pseudo-offspring for the gradient approximation.

Neuroevolution, which is the joint optimization of network topology
and parameters, is another promising application for EAs. This approach
has a long history \cite{Floreano2008} and works well for small networks
up to a few hundred connections. However, scaling this approach to
networks with millions of connections remains a challenge. One recent
line of work \cite{2017arXiv170301041R,Desell:2017:LSE:3067695.3076002,DBLP:journals/corr/abs-1711-00436}
has taken a hybrid approach where the topology is optimized by an
EA but the parameters are still trained with SGD. However, the introduction
or removal of parameters by the EA can be problematic. It may leave
the network in an unfavorable region of the parameter space, with
effects similar to those of a bad initialization at the start of SGD
training. Another line of work has focused on indirect encodings to
reduce the size of the search space \cite{Stanley:2009:HEE:1516090.1516093}.
The difficulty here lies in finding an appropriate mapping from genotype
to phenotype.

\section{Method\label{sec:Method}}

We implement a population-based EA that optimizes the parameters of
directly encoded, fixed size neural networks. For performance reasons,
the EA is implemented with TensorFlow and executes entirely on the
GPU, i.e. the whole population of networks lives in GPU memory and
all EA logic is performed on the GPU.

\subsection{Evolutionary Algorithm}

Algorithm~\ref{alg:ea} shows our EA in pseudo-code. It is a generational
EA extended by the limited evaluation concept. Every generation, the
fitness evaluation is performed on a small batch of data that is drawn
randomly from the training set. This reduces the computational cost
of the fitness evaluation but introduces an increasing amount of noise
with smaller batch sizes. To counteract this, Morse et al. \cite{Morse:2016:SEO:2908812.2908916}
propose a fitness inheritance scheme that we implement as well.

\begin{algorithm}
{
\setstretch{1.25}

\DontPrintSemicolon
\SetProgSty{}

$P \leftarrow \left[ \theta_1, \theta_2, \ldots, \theta_\lambda \,\middle|\, \theta_i \textrm{ randomly initialized} \right] $\;

\While{termination condition not met}{

  $x, y \leftarrow$ select random batch from training data \;
  
  $P \leftarrow P$ sorted by fitness in descending order \;
  
  $E \leftarrow$ select elites $P\left[ \,:\!  p_E \lambda \right]$ \;
  
  $C \leftarrow$ select $p_C \lambda$ parent pairs $\left( \theta_1, \theta_2 \right) \in P\left[ \,:\! \rho \lambda \right]^2$ uniform at random \;
  
  $M \leftarrow$ select $p_M \lambda$ parents $\theta_1 \in P\left[ \,:\!  \rho \lambda \right]$ uniform at random \;
  
  $C' \leftarrow \left[ \textrm{crossover} \left( \theta_1, \theta_2 \right)  \,\middle|\, \left( \theta_1, \theta_2 \right) \in C \right] $ \;
  
  $M' \leftarrow \left[ \textrm{mutation} \left( \theta_1 \right) \,\middle|\, \theta_1 \in M \right] $ \;
  
  $P \leftarrow E \cup C' \cup M'$ \;

  evaluate $\textrm{fitness}\left(\theta,x,y\right)$ for each individual in $\theta \in P$ \;
}

}

\vspace{2mm}

\caption{Evolutionary algorithm. Square brackets indicate ordered lists and
$L\left[\,:\!k\right]$ is notation for the list containing the first
$k$ elements of $L$. \label{alg:ea}}
\end{algorithm}

The initial population is created by randomly initializing the parameters
of $\lambda$ networks. Then, a total of $\lambda$ offspring networks
are derived from the population~$P$. The hyperparameters $p_{E}$,
$p_{C}$ and $p_{M}$ determine the percentage of offspring created
by elite selection, crossover and mutation respectively. First, the
$p_{E}\lambda$ networks with the highest fitness are selected as
elites from the population. These elites move into the next generation
unchanged and will be evaluated again. Even though their parameters
did not change, the repeated evaluation is desirable. Because the
fitness function is only evaluated on a small batch of data, it is
stochastic and repeated evaluations will result in a better estimate
of the true fitness when combined with previous fitness evaluation
results. Next, $p_{C}\lambda$ pairs of networks are selected as parents
for sexual reproduction (crossover) and finally $p_{M}\lambda$ networks
are selected as parents for asexual reproduction (mutation). The selection
procedure in both cases is truncation selection, i.e. parents are
drawn uniform at random from the top $\rho\lambda$ of networks sorted
by fitness, where $\rho\in\left[0,1\right]$ is the selection proportion.

Due to the stochasticity in the fitness evaluation, it seems advantageous
to combine fitness evaluation results from multiple batches. However,
simply evaluating every network on multiple batches is no different
from using a larger batch size. Therefore, the assumption is made
that the fitness of a parent network and its offspring are related.
Then, a parent's fitness can be inherited to its offspring as a good
initial guess and be refined by the actual fitness evaluation of the
offspring. This is done in form of the weighted sum
\[
f_{\textrm{adj}}=\left(1-\alpha\right)\cdot f_{\textrm{inh}}+\alpha\cdot\textrm{fitness}\left(\theta,x,y\right)\,,
\]
where $f_{\textrm{inh}}$ is the fitness value inherited by the parents,
$\textrm{fitness}\left(\theta,x,y\right)$ is the fitness value of
the offspring $\theta$ on the current batch $x,y$ and $\alpha\in\left[0,1\right]$
is a hyperparameter that controls the strength of the fitness inheritance
scheme. Setting $\alpha$ to 1 disables fitness inheritance altogether.
During sexual reproduction of two parents with fitness $f_{1}$ and
$f_{2}$ or during asexual reproduction of a single parent with fitness
$f_{3}$, the inherited fitness values are $f_{\textrm{inh}}=\frac{1}{2}\left(f_{1}+f_{2}\right)$
and $f_{\textrm{inh}}=f_{3}$ respectively.

\subsection{Crossover and Mutation Operators}

Members of the EA population are direct encodings of neural network
parameters $\theta\in\mathbb{R}^{c}$, where $c$ is the total number
of parameters in each network. The crossover and mutation operators
directly modify this vector representation. An explanation of the
crossover and mutation operators that we use in our experiments follows. 

\subsubsection{Uniform crossover.}

The uniform crossover of two parents $\theta_{1}$ and $\theta_{2}$
creates offspring $\theta_{u}$ by randomly deciding which element
of the offspring's parameter vector is taken from which parent:
\[
\theta_{u,i}=\begin{cases}
\theta_{1,i} & \textrm{with probability }0.5\\
\theta_{2,i} & \textrm{else}
\end{cases}
\]

\subsubsection{Arithmetic crossover.}

Arithmetic crossover creates offspring $\theta_{a}$ from two parents
$\theta_{1}$ and $\theta_{2}$ by taking the arithmetic mean:
\[
\theta_{a}=\frac{1}{2}\left(\theta_{1}+\theta_{2}\right)
\]

\subsubsection{Mutation.}

The mutation operator adds random normal noise scaled by a mutation
strength $\sigma$ to a parent $\theta_{1}$: 
\[
\theta_{m}=\theta_{1}+\sigma\cdot\mathcal{N}\left(0,1\right)
\]

The mutation strength $\sigma$ is an important hyperparameter that
can be changed over the course of the EA run if desired. In the simplest
case, the mutation strength stays constant over all generations.

We also experiment with deterministic control in the form of an exponentially
decaying value. For each generation $i$, the mutation strength is
calculated according to $\sigma_{i}=\sigma\cdot0.99^{i/k}$ , where
$\sigma$ is the initial mutation strength and the hyperparameter
$k$ controls the decay rate in terms of generations.

Finally, we implement self-adaptive control. The mutation strength
$\sigma$ is included as a gene in each individual and each individual
is mutated with the $\sigma$ taken from its own genes. The mutation
strength itself is mutated according to $\sigma_{i+1}=\sigma_{i}e^{\tau\mathcal{N}\left(0,1\right)}$
with hyperparameter $\tau$. During crossover, the arithmetic mean
of two $\sigma$-genes produces the value for the $\sigma$-gene in
the offspring.

\subsection{GPU Implementation}

Naively executing thousands of small neural networks on a GPU in parallel
incurs significant overhead, since many short-running, parallel operations
that compete for resources are launched, each of which also has a
startup cost. To efficiently evaluate thousands of network parameter
configurations, the computations should be expressed as batch tensor\footnote{A tensor is a multi-dimensional array.}
products where possible.

Assume we have input data of dimensionality $m$ and want to apply
a fully connected layer with $n$ output units to it. This can naturally
be expressed as a product of a parameter and data tensor with shapes
$\left[n,m\right]\times\left[m\right]=\left[n\right]$, which in this
simple case is just a matrix-vector product. To process a batch of
data at once, a batch dimension $b$ is introduced to the data vector.
The resulting product has shapes $\left[n,m\right]\times\left[b,m\right]=\left[b,n\right]$.
Conceptually, the same product as before is computed for every element
in the data tensor's batch dimension. Batching over multiple sets
of network parameters follows the same approach and introduces a population
dimension $p$. Obviously, the parameter tensor needs to be extended
by this dimension so that it can hold parameters of different networks.
However, the data tensor also needs an additional population dimension
because the output of each layer will be different for networks with
different parameters. The resulting product has shapes $\left[p,n,m\right]\times\left[p,b,m\right]=\left[p,b,n\right]$
and conceptually, the same batch product as before is computed for
every element in the population dimension.

In order to exploit this batched evaluation of populations, the whole
population lives in GPU memory in the required tensor format. Next
to enabling the population batching, this also alleviates the need
to copy data between devices, which reduces latency. These advantages
apply as long as the networks are small enough. The larger each network,
the more computation is necessary to evaluate it, which reduces the
gain from batching multiple networks together. Furthermore, combinations
of population size, network size and batch size are limited by the
available GPU memory. Despite these shortcomings, with 16\,GB GPU
memory this framework allows us to experiment at reasonably large
scales such as a population of 8k networks with 92k parameters each
at a batch size of 64.

\section{Experiments\label{sec:Experiments}}

We apply the EA from Section~\ref{sec:Method} to optimize a neural
network that classifies the MNIST dataset, which is a standard image
classification benchmark with $28\times28$ pixel grayscale inputs
and $d=10$ classes. The training set contains 50k images, which we
split into an actual training set of 45k images and a validation set
of 5k images. All reported accuracies during experiments are validation
set accuracies. The test set of 10k images is only used in the final
experiment that compares the EA to SGD. All experiments have been
repeated 15 times with different random seeds. When significance levels
are mentioned, they have been obtained by performing a one-sided Mann-Whitney-U-Test
between the samples of each experiment. The fitness function to be
maximized by the EA is defined as the negative, average cross-entropy
\begin{equation}
-\frac{1}{n}\sum_{i=1}^{n}H\left(p_{i},q_{i}\right)=\frac{1}{nd}\sum_{i=1}^{n}\sum_{j=1}^{d}p_{ij}\log\left(q_{ij}\right)\,,\label{eq:fitnessfunction}
\end{equation}
where $n$ is the batch size, $p_{ij}\in\left\{ 0,1\right\} $ is
the ground-truth probability and $q_{ij}\in\left[0,1\right]$ is the
predicted probability for the $j$th class in the $i$th example.
Unless otherwise stated, the following hyperparameters are used for
experiments:
\[
\begin{aligned}\textrm{crossover op.} & =\textrm{uniform} & p_{E} & =0.05 & \lambda & =1000 & \alpha & =1.00\\
\textrm{sigma adapt.} & =\textrm{constant} & p_{C} & =0.50 & \sigma & =0.001\\
\textrm{batch size} & =512 & p_{M} & =0.45 & \rho & =0.50
\end{aligned}
\]

\subsection{Neural Network Description}

The neural network we use in all our experiments applies $2\times2$
max-pooling to its inputs, followed by four fully connected layers
with 256, 128, 64 and 10 units respectively. Each layer except for
the last one is followed by a ReLU non-linearity. Finally, the softmax
function is applied to the network output. In total, this network
has 92k parameters that need to be trained. 

This network is unable to achieve state-of-the-art results even with
SGD training but has been chosen due to the following considerations.
We wanted to limit the maximum network parameter count to roughly
100k so that it remains possible to experiment with large populations
and batch sizes. However, we also wanted to work with a multi-layer
network. We deem this aspect important, as there should be additional
difficulty in optimizing deeper networks with more interactions between
parameters. To avoid concentrating a large part of the parameters
in the network's first layer, we downsample the input. This way, it
is possible to have a multi-layer network with a significant number
of parameters in all layers. Furthermore, we decided against using
convolutional layers as our batched implementation of fully connected
layers is more efficient than the convolutional counterpart.

All networks for the EA population are initialized using the Glorot-uniform~\cite{pmlr-v9-glorot10a}
initialization scheme. Even though Glorot-uniform and other neural
network initialization schemes were devised to improved SGD performance,
we find that the EA also benefits from them. Furthermore, this allows
for a comparison to SGD on even footing.

\subsection{Tradeoff between Batch Size and Accuracy}

\begin{figure}
\begin{centering}
\includegraphics[width=0.83\textwidth]{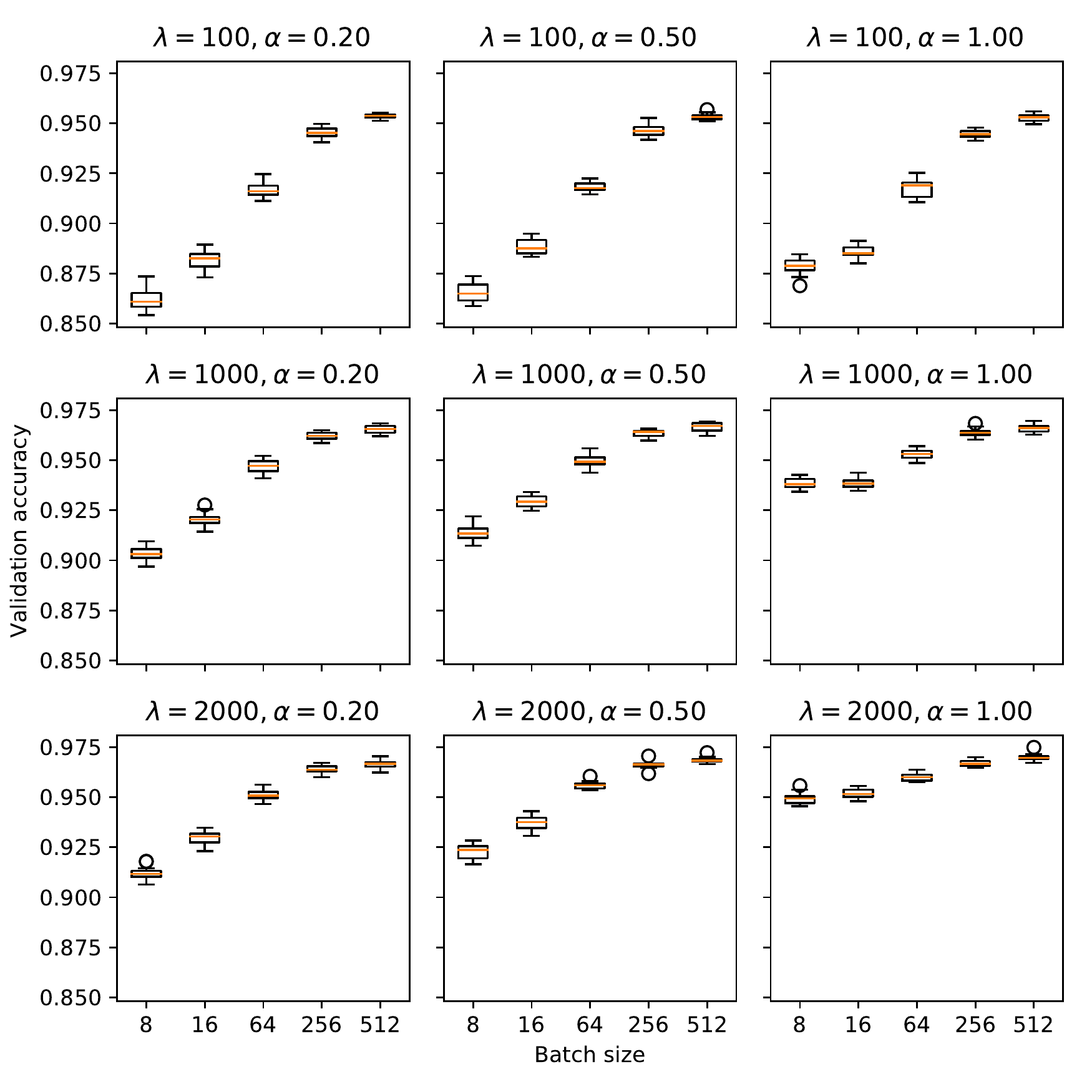}
\par\end{centering}
\caption{Validation accuracies of 15 EA runs for different population sizes
$\lambda$, fitness inheritance strengths $\alpha$ and batch sizes.
Looking at the grid of figures, $\lambda$ increases from top to bottom,
while $\alpha$ increases from left to right. A box extends from the
lower to upper quartile values of the data, with a line at the median
and whiskers that show the range of the data. \label{fig:lmbda-batch-alpha}}
\end{figure}

The EA chooses a batch of training data for each generation and uses
it to evaluate the population's fitness. A single fitness evaluation
is therefore only a noisy estimate of the true fitness. The smaller
the batch size, the noisier this estimate becomes because Equation~\ref{eq:fitnessfunction}
averages over fewer cross-entropy loss values. A noisy fitness estimate
introduces two problems: A good network may receive a low fitness
value and be eliminated during selection or a bad network may receive
a high fitness value and survive. The fitness inheritance was introduced
by Morse et al. \cite{Morse:2016:SEO:2908812.2908916} with the intent
to counteract this noise and allow effective optimization despite
noisy fitness values. However, in preliminary experiments fitness
inheritance did not seem to have a positive impact on our results,
so we performed a systematic experiment to explore the interaction
between batch size, fitness inheritance and the resulting network
accuracy. The results can be found in Figure~\ref{fig:lmbda-batch-alpha}.
Three key observations can be made:

First of all, the validation set accuracy is positively correlated
with the batch size. This relationship holds for all tested settings
of $\lambda$ and $\alpha$. This means, using larger batch sizes
gives better results. Note that the EA was allowed to run for more
generations when the batch size was small, so that all runs could
converge. In consequence, it is not possible to compensate the accuracy
loss incurred by small batch sizes by allowing the EA to perform more
iterations.

Second, the validation set accuracy is also positively correlated
with $\alpha$. Especially for small batch sizes, significant increases
in validation accuracy can be observed when increasing $\alpha$.
This is surprising as higher values of $\alpha$ reduce the amount
of fitness inheritance. Instead, we find that the fitness inheritance
either has a harmful or no effect.

Lastly, increasing the population size $\lambda$ improves the validation
accuracy. This is important but unsurprising as increasing the population
size is a known way to counteract noise \cite{Beyer98evolutionaryalgorithms}.

\subsection{Selective Pressure}

Having observed that fitness inheritance does not improve results
at small batch sizes, we will now show that instead decreasing the
selective pressure helps. The selective pressure influences to what
degree fitter individuals are favored over less fit individuals during
the selection process. Since small batches produce noisy fitness evaluations,
a low selective pressure should be helpful because the EA is less
likely to eliminate all good solutions based on inaccurate fitness
estimates.

We experiment with different settings of the selection proportion
$\rho$, which determines what percentage of the population ordered
by fitness is eligible for reproduction. During selection, parents
are drawn uniformly at random from this group. Low selection proportions
(low values of $\rho$) lead to high selective pressure because parents
are drawn from a smaller group of individuals with high (apparent)
fitness. Therefore, we expect high values of $\rho$ to work better
with small batches.

\begin{figure}
\begin{centering}
\includegraphics[width=0.65\textwidth]{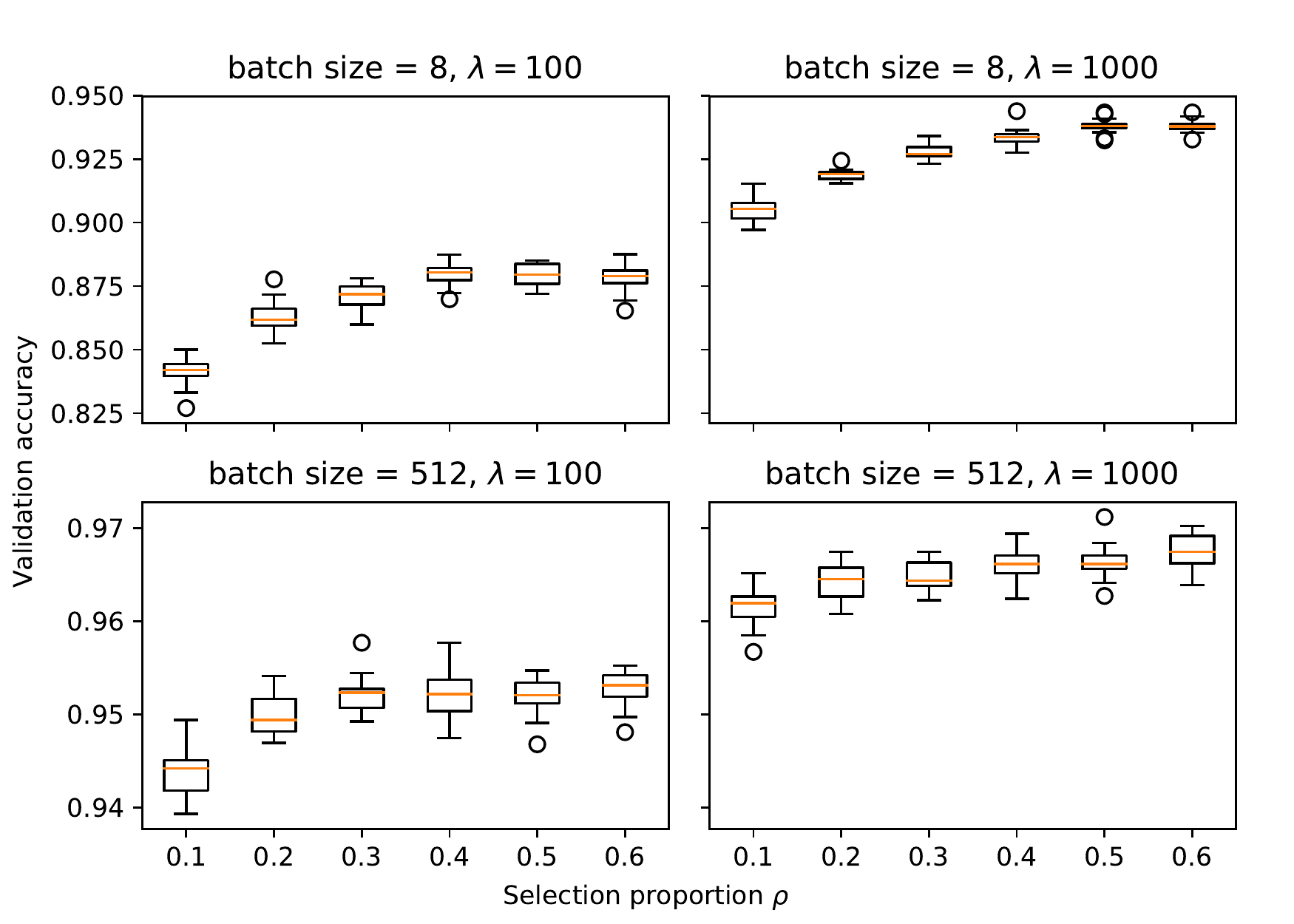}
\par\end{centering}
\caption{Validation accuracies of 15 EA runs for different population sizes
$\lambda$, batch sizes and selection proportions $\rho$. The first
row of figures shows results for small batch sizes, while the second
row shows results for large batch sizes. \label{fig:lmbda-batch-trunc}}
\end{figure}

Figure~\ref{fig:lmbda-batch-trunc} shows results for increasing
values of $\rho$ at two different batch sizes and two different population
sizes. Generally speaking, increasing $\rho$ increases the validation
accuracy (up to a certain degree). For a specific $\rho$ it is unfortunately
not possible to compare validation accuracies across the four scenarios,
because batch size and population size are influencing factors as
well. Instead, we treat the relative difference in validation accuracies
going from $\rho=0.1$ to $\rho=0.2$ as a proxy. Table~\ref{tab:selective-pressure}
confirms that decreasing the selective pressure (by increasing $\rho$)
has a positive influence on the validation accuracy.

\begin{table}
\small
\setlength{\tabcolsep}{10pt}

\caption{Relative improvement in validation accuracy when increasing the selection
proportion from $\rho=0.1$ to $\rho=0.2$ in four different scenarios.
Since large population sizes are also an effective countermeasure
against noise, the relative improvement decreases with increasing
population sizes. The fitness noise column only depends on batch size
and is included to highlight the correlation between noise and relative
improvement. \label{tab:selective-pressure}}
\centering{}%
\begin{tabular}{lllc}
\toprule 
Batch size & Fitness noise & Population size & Relative improvement\tabularnewline
\midrule
\midrule 
8 & high & 100 & 2.26\,\%\tabularnewline
8 & high & 1000 & 1.57\,\%\tabularnewline
512 & low & 100 & 0.49\,\%\tabularnewline
512 & low & 1000 & 0.34\,\%\tabularnewline
\bottomrule
\end{tabular}
\end{table}

\subsection{Crossover and Mutation Operators}

While the previous experiments explored the influence of limited evaluation,
another significant factor for good performance are crossover and
mutation operators that match the optimization problem. Neural networks
in particular have problematic redundancy in their search space: Nodes
in the network can be reordered without changing the network connectivity.
This means, there are multiple equivalent parameter vectors that represent
the same function mapping.

\begin{figure}
\begin{centering}
\includegraphics[width=0.6\textwidth]{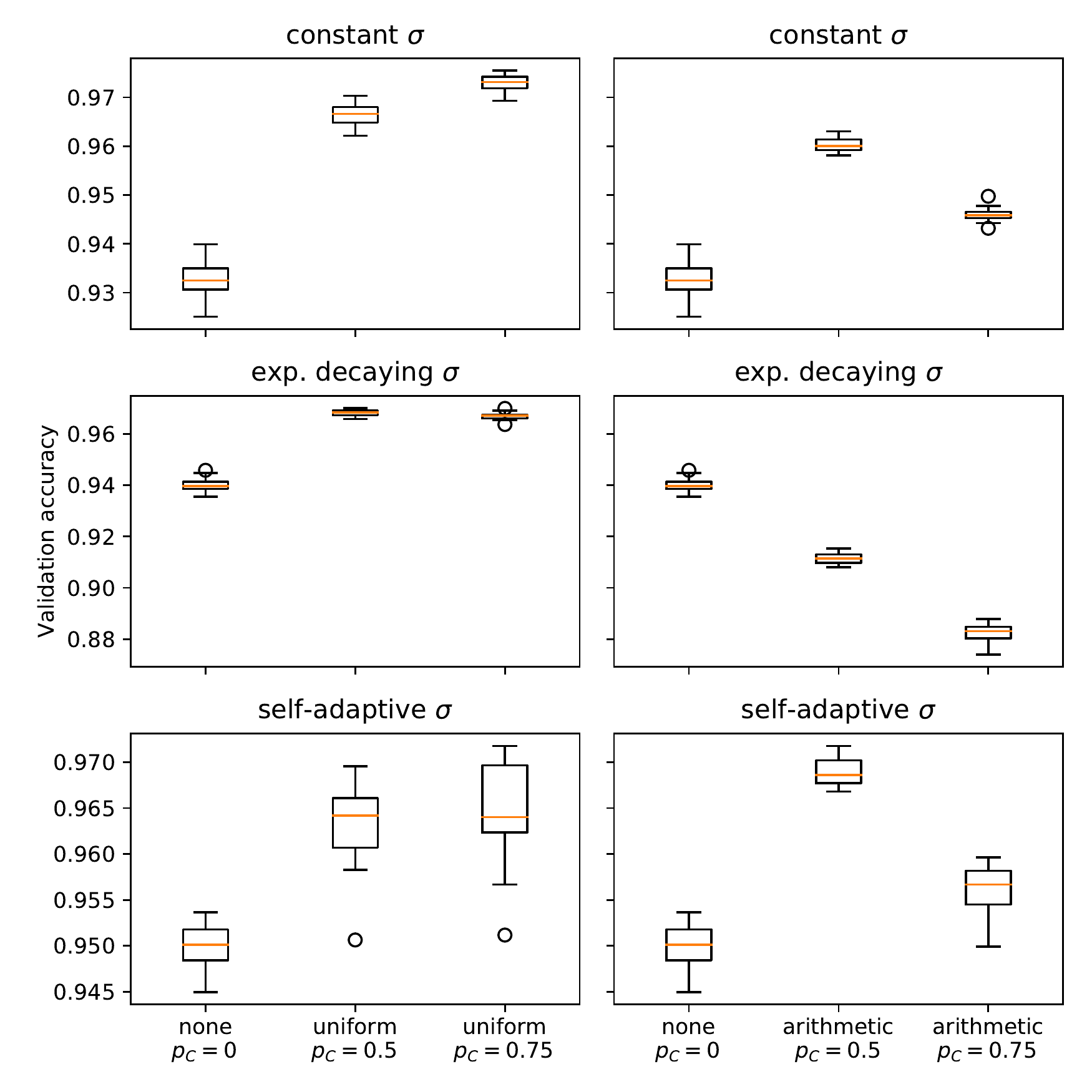}
\par\end{centering}
\caption{Validation accuracies of 15 EA runs with different levels of crossover
$p_{C}$, crossover operators and mutation strength $\sigma$ adaptation
schemes. The left column shows results using uniform crossover, while
arithmetic crossover is employed for the right column. \label{fig:xo-mut}}
\end{figure}

Designing crossover and mutation operators that are specifically equipped
to deal with these problems seems like a promising research direction,
but for now we want to establish baselines with commonly used operators.
In particular, these are uniform and arithmetic crossover as well
as random normal mutation. It is not obvious if crossover is helpful
for optimizing neural networks as there is no clear compositionality
in the parameter space. There are many interdependencies between parameters
that might be destroyed, e.g. when random parameters are replaced
by those from another network during uniform crossover. Therefore,
we not only want to compare the uniform and arithmetic crossover operators
among themselves, but also test if crossover leads to improvements
at all. This can be achieved by varying the EA hyperparameter $p_{C}$
, which controls the percentage of offspring that are created by the
crossover operator. On the other hand, random normal mutation intuitively
performs the role of a local search but its usefulness significantly
depends on the choice of the mutation strength $\sigma$. Therefore,
we compare three different adaptation schemes: constant, exponential
decay and self-adaptation.

\begin{figure}
\begin{centering}
\includegraphics[width=0.8\textwidth]{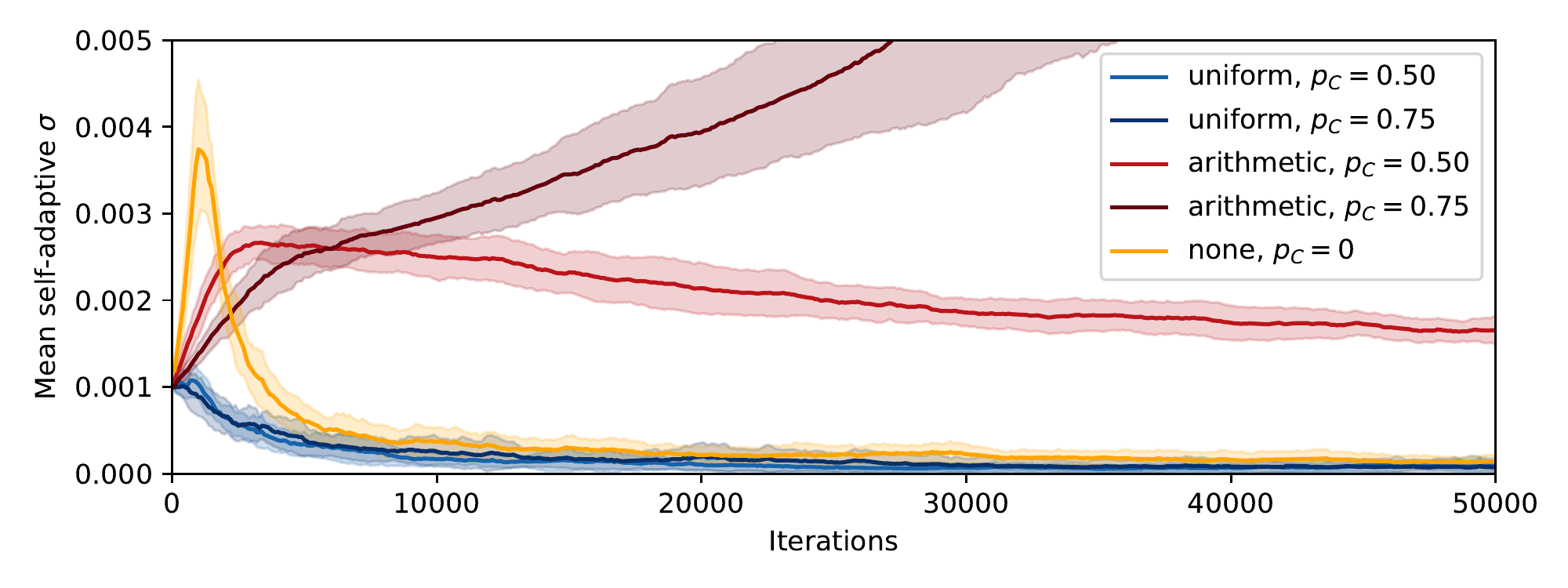}
\par\end{centering}
\caption{Population mean of $\sigma$ from 15 EA runs with self-adaptation
turned on. The shaded areas indicate one standard deviation around
the mean. \label{fig:sigma-selfadaptive}}
\end{figure}

Since crossover operators might need different mutation strengths
to operate optimally, we test all combinations and show results in
Figure~\ref{fig:xo-mut}. Using crossover ($p_{C}>0$) always results
in significantly ($p<0.01$) higher validation accuracy than not using
crossover ($p_{C}=0$), except for the case of arithmetic crossover
with exponential decay. The reason for this is likely, that arithmetic
crossover needs high mutation strengths but the exponential decay
decreases $\sigma$ too fast. This becomes evident when examining
the mutation strengths chosen by self-adaptation in Figure~\ref{fig:sigma-selfadaptive}.
Compared to uniform crossover, the self-adaptation drives $\sigma$
to much higher values when arithmetic crossover is used. Overall,
both crossover operators work well under different circumstances.
Uniform crossover at $p_{C}=0.75$ with constant $\sigma$ achieves
the highest median validation accuracy of 97.3\,\%, followed by arithmetic
crossover at $p_{C}=0.5$ with self-adaptive $\sigma$ at 96.9\,\%
validation accuracy. When using uniform crossover at $p_{C}=0.75$,
a constant mutation strength works significantly ($p<0.01$) better
than the other adaptation schemes. On the other hand, for arithmetic
crossover at $p_{C}=0.5$, the self-adaptive mutation strength performs
significantly ($p<0.01$) better than the other two tested adaptation
schemes. The main drawback of the self-adaptive mutation strength
is the the additional randomness that leads to high variance in the
training results.

\subsection{Comparison to SGD}

Informed by the other experiments, we want to run the EA with advantageous
hyperparameter settings and compare its test set performance to the
Adam optimizer. Most importantly, we use a large population, large
batch size, no fitness inheritance, and offspring are created by uniform
crossover in 75\,\% of all cases:
\[
\begin{aligned}\textrm{crossover op.} & =\textrm{uniform} & p_{E} & =0.05 & \lambda & =2000 & \alpha & =1.00\\
\textrm{sigma adapt.} & =\textrm{constant} & p_{C} & =0.75 & \sigma & =0.001\\
\textrm{batch size} & =1024 & p_{M} & =0.20 & \rho & =0.50
\end{aligned}
\]
Median test accuracies over 15 repetitions are 97.6\,\% for the EA
and 98.0\,\% for Adam. Adam still significantly ($p<0.01$) beats
EA performance, but the difference in final test accuracy is rather
small. However, training with Adam progresses about 10 times faster
so it would be wrong to claim that EAs are competitive for neural
network training. Yet, this work is another piece of evidence that
EAs have potential for applications in this domain.

\section{Conclusion\label{sec:Conclusion}}

Efficient batch fitness evaluation of a population of neural networks
on GPUs made it feasible to perform extensive experiments with the
LEEA. While the idea of using very small batches for fitness evaluation
is appealing for computational cost reasons, we find that it comes
with the drawback of significantly lower accuracy than with larger
batches. Furthermore, the fitness inheritance that is supposed to
offset such drawbacks actually has a detrimental effect in our experiments.
Instead, we propose to use low selective pressure as an alternative.

We compare uniform and arithmetic crossover in combination with different
mutation strength adaptation schemes. Surprisingly, uniform crossover
works best among all tested combinations even though it is counter-intuitive
that randomly replacing parts of a network's parameters with those
of another network is helpful.

Finally, we train a network of 92k parameters on MNIST using an EA
and reach an average test accuracy of 97.6\,\%. SGD still achieves
higher accuracy at 98\,\% and is remarkably more efficient in doing
so. However, having demonstrated that EAs are able to optimize large
neural networks, future work may focus on the application to areas
such as neuroevolution where EAs may have a bigger edge.

\bibliographystyle{splncs04}
\bibliography{gpuea}

\end{document}